\documentclass{IET}

\begin{document}

\title{An Adaptive Psychoacoustic Model for Automatic Speech Recognition}

\author[1,2, *]{Peng Dai}
\affil{Department of Computer Science, University of Western Ontario, London, ON}

\author{Xue Teng}
\affil{Pulse Infoframe Inc., London, ON}

\author{Frank Rudzicz}
\affil{Toronto Rehabilitation Institute - UHN, Toronto, ON, Canada}

\author{Ing Yann Soon}
\affil{School of Electrical and Electronic Engineering, Nanyang Technological University, Singapore}
\affil[*]{peng.dai.ca@ieee.org}

\abstract{Compared with automatic speech recognition (ASR), the human auditory system is more adept at handling noise-adverse situations, including environmental noise and channel distortion. To mimic this adeptness, auditory models have been widely incorporated in ASR systems to improve their robustness. This paper proposes a novel auditory model which incorporates psychoacoustics and otoacoustic emissions (OAEs) into ASR. In particular, we successfully implement the frequency-dependent property of psychoacoustic models and effectively improve resulting system performance. We also present a novel double-transform spectrum-analysis technique, which can qualitatively predict ASR performance for different noise types. Detailed theoretical analysis is provided to show the effectiveness of the proposed algorithm.
Experiments are carried out on the AURORA2 database and show that the word recognition rate using our proposed feature extraction method is significantly increased over the baseline. Given models trained with clean speech, our proposed method achieves up to 85.39\% word recognition accuracy on noisy data.
}

\maketitle

\section{Introduction}
Speech may be the most important form human communication, and automatic speech recognition (ASR) has received considerable attention as a result. After decades of development, ASR has become very effective in decoding clean speech, e.g., achieving over 95\% word accuracy in small vocabulary contexts and over 90\% in large vocabulary contexts given speech wih signal-to-noise ratios above 20~dB \cite{Ramirez2011,B99Gold}. However, as SNR drops (e.g., to 0 dB), the recognition accuracy can fall below 50\%, which is not acceptable for many typical applications.  This is in contrast to the human auditory system, which shows greater resilience to noise \cite{rabiner,Virag1999}. For humans, speech perception is a sensory and perceptual process  \cite{Dai12sc,cit_tm1,B99Gold} and in this paper we focus on the psychoacoustic and otoacoustic emission (OAE) aspects of that process.


Psychoacoustics is the broad investigation of human speech perception and includes relationships between sound pressure level and loudness, human response to different frequencies, and a variety of masking effects \cite{B99Gold,Dai12sc}. To some extent, the popularity of Mel-frequency cepstral coefficients (MFCCs) are a result of this area of research \cite{Davis80,Milner02}. Otoacoustic emissions (OAEs) are acoustic signals produced in the cochlea, which is widely used in the diagonisis of hearing loss for newborns \cite{handbook08} but have not really been applied in ASR. When the cochlea is stimulated by external acoustic signals, the outer hair cells vibrate, which produces a nearly inaudible sound that echoes back into the middle ear \cite{handbook08}.

Our previous work in psychoacoustics systematically investigated how speech signals are processed by the human auditory system and converted to neural spikes \cite{Dai10sc,Dai12sc,Dai09ICICS}. In particular, we proposed several different mathematical models for the effective implementation of masking effects, which describe the phenomenon that a clearly audible sound (maskee) becomes weak or inaudible in the presence of another sound (masker). We have also improved aspects of ASR by incorporating temporal integration \cite{Oxenham00,Oxenham01}.

In this paper, we further improve the auditory model. Our major contributions consist of three parts. First, we successfully implement the frequency-dependent property of masking effects. Moreover, we propose an approximation for OAEs, which is incorporated into the ASR system. Finally, we present novel theoretical and quantitative justifications for this incorporation. In particular, we propose a novel analysis technique which can be used to predict the ASR performance for different noise types and algorithms. 

\subsection{Auditory model}\label{sc:hum}
In this work, we study two subareas of auditory neuroscience, namely psychoacoustics and otoacoustic emissions (OAEs). Psychoacoustics covers many different topics, including limits of perception, sound localization, and masking effects. The masking effect is the phenomenon in which a clearly audible sound (maskee) is influenced by another sound (masker). To measure the effect of masking quantitatively, a masking threshold is usually determined. The masking threshold is the sound pressure level of a test sound, to be barely audible in the presence of a masker. Masking effects may be classified as simultaneous or temporal according to signal occurrence \cite{Dai12sc}. Masking effects between any two signals which occur at the same time is {\em simultaneous} or {\em frequency} masking. Signals can be masked by the preceding sound, called {\em forward} masking, or by the subsequent sound, called {\em backward} masking. Temporal masking can be viewed as a consequence of auditory adaptation \cite{Strope97}. These masking effects are caused by the principal mechanism of neuronal signal processing in both time and frequency  \cite{ShammaII85,ShammaI85,Kvale04}.

Otoacoustic emissions (OAE) are acoustic signals generated from within the inner ear, which can be recorded in the ear canal using a sensitive microphone \cite{handbook08}. Otoacoustic emissions (OAE) are a consequence of the nonlinear and active pre-processing of sound in the cochlea \cite{handbook08}. Predicted by Thomas Gold in 1948, OAE was first demonstrated empirically by David Kemp in 1978 \cite{Kemp1978} and otoacoustic emissions have since been shown to arise through a number of different cellular and mechanical causes within the inner ear \cite{chang97,Kujawa1996}. Studies have shown that OAEs disappear after the inner ear has been damaged, so OAEs are often used in the laboratory and clinic as a measure of inner ear health \cite{handbook08}. 

The organization of this paper is as follows. Detailed derivations and algorithm descriptions are given in Section \ref{sc:2Dlow_algorithm}. This is followed by the theoretical anlysis of the noise reduction ability of the proposed algorithm and a novel double transform domain analysis technique in Section \ref{sc:2D_theoretical}. The experimental databases and detailed settings are given in Section \ref{sc:2D_result}. Finally, we conclude our work in Section \ref{sc:2D_conclusion}.


\section{Algorithm Description}\label{sc:2Dlow_algorithm}
In this part, we will describe our proposed mathematical model for the human auditory system. It mainly consists of two parts, adaptive 2D psychoacoustic filter and the OAE filter.

\subsection{2D psychoacoustic filter}\label{ssc:masking}


 Forward masking (FM) reveals that over short durations, the usable dynamic range of the human auditory system depends on the spectral characteristics of the previous stimuli \cite{Dai10sc}. Backward masking describes how a speech signal is affected by subsequent stimuli. A masking threshold is usually defined to describe the extent to which the masker affects the maskee. Since masking effects modify both the time and frequency components of acoustic signals, our proposed algorithm is designed in the joint time-frequency domain.

A speech signal, $y(t)$, is split into frames and transformed to the time-frequency domain, represented as $Y(f,t)$, by the Fourier transform. Here, $f$ and $t$ are frequency (band) and time (frame) indices of the signal, respectively. Since $f$ and $t$ can be converted to the actual frequency and time of the signal, for simplicity they are used interchangably as the actual frequency and time in the following disussion.



Temporal masking can be modeled as
\begin{equation}\label{eq:me1}
M_{tm}(f,t,\Delta t)=A_{tm}(\Delta t)Y(f,t+\Delta t),
\end {equation}
where $A_{tm}(f,\Delta t)$ is the temporal masking parameter given in \cite{Dai12sc}; $M_{tm}$ is the amount of temporal masking; and $\Delta t$ is the signal delay \cite{Dai12sc,Oxenham00,Jesteadt82}. Equation \eqref{eq:me1} describes how a speech signal, $Y(f,t+\Delta t)$, can affect other acoustic signals that occur at different times. Similarly, simultaneous masking can be modeled as Equation \eqref{eq:me2}, and temporal-frequency masking can be modeled as Equation \eqref{eq:me3} \cite{Dai12sc}.
\begin{equation}\label{eq:me2}
M_{sm}(f,t,\Delta f)=A_{sm}(\Delta f)Y(f+\Delta f,t)
\end {equation}
\begin{equation}\label{eq:me3}
M_{diag}(f,t,\Delta f,\Delta t)=A_{diag}(\Delta f,\Delta
t)Y(f+\Delta f,t+\Delta t)
\end {equation}

In the time-frequency domain, speech components are influenced by nearby surrounding components. In other words, a speech signal, $Y(f,t)$, is affected by all other speech signals within a certain range,
$\{\: Y(f+\Delta f,t+\Delta t)\:|\:\: -T_{bm}\leq \Delta t \leq T_{fm},-F_{1}\leq \Delta f \leq F_{2} \}$. $T_{fm}$ and $T_{bm}$ are the effective ranges of forward masking and backward masking, respectively, and $F_1$ and $F_2$ are the effective range of simultaneous masking.

The overall joint masking effect can be described as
\begin{eqnarray}
\label{eq:der25}
  &&\ M_{total}(f,t)  \nonumber \\
  &=& \sum_{\Delta t =-T_{bm}}^{T_{tm}}A_{tm}(\Delta t)Y(f,t+\Delta t) \nonumber \\
  && +\sum_{\Delta f \neq 0}A_{sm}(\Delta f)Y(f+\Delta f,t)  \nonumber \\
  && +\sum_{\Delta t \neq 0}\sum_{\Delta f \neq 0}A_{diag}(\Delta f,\Delta t)Y(f+\Delta f,t+\Delta t).
\end{eqnarray}
Then, the total masking effect becomes
\begin{eqnarray}\label{eq:der27}
  &&\ M_{total} \nonumber  \\
  &=& {\sum_{\Delta t=-T_{bm}}^{T_{fm}}\sum_{\Delta f=-F_1}^{F_2}\alpha(\Delta f,\Delta t)Y(f+\Delta f,t+\Delta t)}, \nonumber\\
\end{eqnarray}
where $\alpha(\Delta f,\Delta t)$ is the filter parameter, defined by
\begin{equation}\label{eq:der26}
\alpha(\Delta f,\Delta t)=
\begin{cases}
0                               	& \Delta f = 0,\Delta f = 0\\
A_{tm}(\Delta t)              	& \Delta f = 0,\Delta t \neq 0\\
A_{sm}(\Delta f)              	& \Delta f \neq 0,\Delta t = 0\\
A_{diag}(\Delta f,\Delta t)	& \Delta f \neq 0,\Delta t \neq 0
\end{cases}
\end{equation}

\begin{equation}\label{eq:2dfilter}
\begin{array}{l}
 {\bf{ Mask }}
  = \left[ {\begin{array}{*{20}{c}}
   {{{\bf{0}}_{({F_1} - {F_2}) \times T}}} & {{{\bf{0}}_{ ({F_2} - {F_1}) \times T+1}}}  \\
   {{{\bf{0}}_{({F_1} + {F_2} + 1) \times T}}} & {\begin{array}{*{20}{c}}
   { - \alpha \left( {{F_2},0} \right)} & {} & {} & { - \alpha \left( {{F_2}, - T_{fm}} \right)}  \\
    \vdots  & {} &  {\mathinner{\mkern2mu\raise1pt\hbox{.}\mkern2mu
 \raise4pt\hbox{.}\mkern2mu\raise7pt\hbox{.}\mkern1mu}}  & {}  \\
   { - \alpha \left( {1,0} \right)} & { - \alpha \left( {1, - 1} \right)} & {} & {}  \\
   {1  } & { - \alpha \left( {0, - 1} \right)} &  \cdots  & { - \alpha \left( {0, - T_{fm}} \right)}  \\
   { - \alpha \left( { - 1,0} \right)} & { - \alpha \left( { - 1, - 1} \right)} & {} & {}  \\
    \vdots  & {} &  \ddots  & {}  \\
   { - \alpha \left( { - {F_1},0} \right)} & {} & {} & { - \alpha \left( { - {F_1}, - T_{fm}} \right)}  \\
\end{array}}  \\
\end{array}} \right] \\
 \end{array}
\end{equation}

\begin{equation}\label{eq:der36}
\begin{array}{l}
 {\bf{{\hat M}}}
  = \left[ {\begin{array}{*{20}{c}}
   {{{\bf{0}}_{({F_1} - {F_2}) \times T}}} & {{{\bf{0}}_{ ({F_2} - {F_1}) \times T+1}}}  \\
   {{{\bf{0}}_{({F_1} + {F_2} + 1) \times T}}} & {\begin{array}{*{20}{c}}
   { - \alpha \left( {{F_2},0} \right)} & {} & {} & { - \alpha \left( {{F_2}, - T_{fm}} \right)}  \\
    \vdots  & {} &  {\mathinner{\mkern2mu\raise1pt\hbox{.}\mkern2mu
 \raise4pt\hbox{.}\mkern2mu\raise7pt\hbox{.}\mkern1mu}}  & {}  \\
   { - \alpha \left( {1,0} \right)} & { - \alpha \left( {1, - 1} \right)} & {} & {}  \\
   {1 + {\alpha _{TI}}} & { - \alpha \left( {0, - 1} \right)} &  \cdots  & { - \alpha \left( {0, - T_{fm}} \right)}  \\
   { - \alpha \left( { - 1,0} \right)} & { - \alpha \left( { - 1, - 1} \right)} & {} & {}  \\
    \vdots  & {} &  \ddots  & {}  \\
   { - \alpha \left( { - {F_1},0} \right)} & {} & {} & { - \alpha \left( { - {F_1}, - T_{fm}} \right)}  \\
\end{array}}  \\
\end{array}} \right] \\
 \end{array}
\end{equation}

The masked speech that, in theory, is transmitted on the auditory nerves to the human brain can then be expressed as
\begin{eqnarray}\label{eq:alg1}
  &&\ \tilde{Y}(f,t) \nonumber \\
  &=& Y(f,t)-M_{total} \\
  &=& Y(f,t)\otimes \bf{Mask} \nonumber
\end{eqnarray}
where $\bf{Mask}$ is defined in Equation \eqref{eq:2dfilter} \cite{Dai10sc,Dai12sc,Dai09ICICS}. Because backward masking is relatively weak compared with forward masking, only forward masking is included in the 2D psychoacoustic filter.

\begin{figure}[htb]
    \centering
    \includegraphics[width=0.45\textwidth]{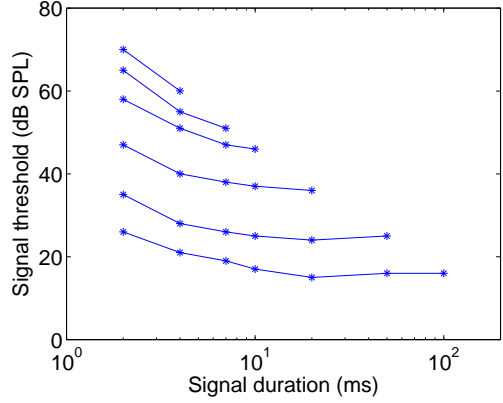}
    \caption{Temporal integration results, from \cite{Oxenham01}. }
    \label{fig:TIcurve}
\end{figure}

Masking effects are generally described in terms of their temporal and frequency aspects. However, the duration of speech signals can also greatly affect the total masking, which is called {\em temporal integration} (TI). According to \cite{Oxenham00,Oxenham01}, when signal durations increase, there is a considerable decrease in the mean masking thresholds (or the amount of masking). For example, Figure \ref{fig:TIcurve} (from \cite{Oxenham01}: Fig 1, pp735), shows that at an offset of 9 ms, mean thresholds decreased by nearly 14 dB as the signal duration increased from 2 to 7 ms. In other words, in Oxenham's experiment, an increase of 5 ms (7 ms - 2 ms) in signal lengths resulted in a 14-dB decrease in the amount of masking. Note that at the duration of 2 ms, the amount of masking is about 56 dB. Notably, the amount of masking drops by about 25\% due to a slight increase (5 ms) in the signal duration.

Since speech has active/non-active periods, its power is more concentrated at certain time, both stronger in energy and longer in duration.
Therefore, temporal integration tends to greatly influence perceived speech. The total masking then becomes
\begin{equation}\label{eq:der33}
M_{psy}^{} = \left\{ {
\begin{array}{*{20}{c}}
   {M_{total} - M^{TI}_1} & {,nonspeech}  \\
   {M_{total} - M^{TI}_2} & {,speech}  \\
\end{array}}
\right.
\end{equation}
where $M^{TI}_1$ and $M^{TI}_2$ are the decreases of masking caused by temporal integration, and $M^{TI}_1 < M^{TI}_2$. Then,
\begin{eqnarray}\label{eq:der35}
  &&\ \tilde{Y}(f,t) \nonumber \\
  &=& Y(f,t)-M_{psy} \\
  &=& \left\{ {
\begin{array}{*{20}{c}}
   {Y\left( {f,t} \right) - M_{total} + {M^{TI}_1}} & {,nonspeech}  \\
   {Y\left( {f,t} \right) - M_{total} + {M^{TI}_2}} & {,speech}  \\
\end{array}}
\right. \nonumber.
\end{eqnarray}

In our present implementation, temporal integration is calculated by
\begin{equation}\label{eq:der34}
M_{TI} = \alpha_{TI}Y(f_i,t_i)
\end{equation}
where $\alpha_{TI}$ is the parameter for calculating TI. It has to be noted that $\alpha_{TI}$ takes different values for different conditions.

The 2D psychoacoustic filter is therefore
\begin{equation}\label{eq:der36a}
{\bf{Mask}} = \left[ {\begin{array}{*{20}{c}}
   {{{\bf{0}}_{\left( {{F_1} - {F_2}} \right) \times T_{fm}}}} & {{{\bf{0}}_{\left( {{F_2} - {F_1}} \right) \times \left( {T_{fm} + 1} \right)}}}  \\
   {{{\bf{0}}_{\left( {{F_1} + {F_2} + 1} \right) \times T_{fm}}}} & {\hat M}  \\
\end{array}} \right],
\end{equation}
where ${\bf{{\hat M}}}\left( f \right)$ is defined in Equation \eqref{eq:der36}.

The proposed 2D psychoacoustic filter enhances the high frequencies
and helps to sharpen the spectral peaks so as to improve the
performance of the ASR system. For simplicity, $\hat M$ will hereafter
be referred to as the 2D psychoacoustic filter.

\subsection{Adaptive 2D Psychoacoustic Filter}\label{ssc:a2D tfw}
\begin{figure}[htb]
    \centering
    \includegraphics[width=0.45\textwidth]{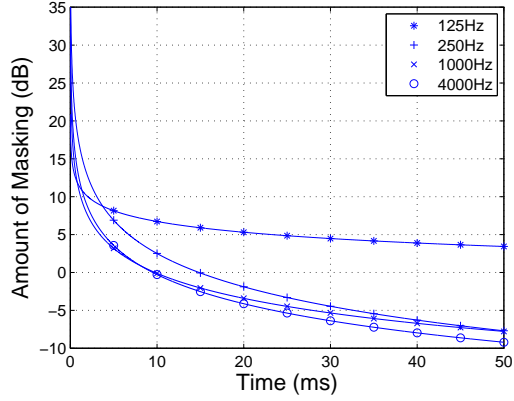}
    \vspace{-5pt}
    \caption{Characteristic curve of forward masking \cite{Jesteadt82,Dai12sc}}
    \label{fig:FMcurve}
\end{figure}

The human auditory system responds differently to different frequencies and masking effects are likewise frequency-dependent. That is, the frequency of the masker affects the total amount of masking, $M_{total}$, which means the parameter $\alpha(\Delta f, \Delta t)$ (see Equation \eqref{eq:der36}) changes with frequency. Figure \ref{fig:FMcurve} shows the characteristic curve of forward masking, which describes how the amount of masking, ${\bf M_{total}}$, changes with time, $\Delta t$ \cite{Jesteadt82}. The 1 kHz and 4 kHz parameters are used for low-band and high-band temporal masking parameters, respectively.

As the parameters of masking effects change with frequency, ideally there should be different 2D psychoacoustic filters for different frequencies, but this can be impractical computationally. Therefore, in our present implementation, we divide each speech sample, denoted as $F_s \times T_s$ matrix ${\bf Y}_s$, into two parts, namely the low and high frequency bands.
\begin{equation}\label{eq:a2D1}
{\bf{Y}_s} = \left[ {\begin{array}{*{20}{c}}
   {{ {\bf Y}_{s1} }}  \\
   {{ {\bf Y}_{s2} }}  \\
\end{array}} \right]
\end{equation}
where ${\bf{Y}}_{{{s1}}}$ and ${\bf{Y}}_{s2}$ are defined as
\begin{equation}\label{eq:a2D2}
{{\bf{Y}}_{{{s1}}}} = \left[ {\begin{array}{*{20}{c}}
   {Y\left( {1,1} \right)} & {Y\left( {1,2} \right)} &  \cdots  & {Y\left( {1,{T_s}} \right)}  \\
    \vdots  &  \vdots  & {} &  \vdots   \\
   {Y\left( {\frac{F_s}{2},1} \right)} & {Y\left( {\frac{F_s}{2} + 1,2} \right)} &  \cdots  & {Y\left( {\frac{F_s}{2} + 1,{T_s}} \right)}  \\
\end{array}} \right]
\end{equation}

\begin{equation}\label{eq:a2D3}
{{\bf{Y}}_{{{s2}}}} = \left[ {\begin{array}{*{20}{c}}
   {Y\left( {\frac{F_s}{2} + 1,1} \right)} & {Y\left( {\frac{F_s}{2} + 1,2} \right)} &  \cdots  & {Y\left( {\frac{F_s}{2} + 1,{T_s}} \right)}  \\
    \vdots  &  \vdots  & {} &  \vdots   \\
   {Y\left( {{F_s},1} \right)} & {Y\left( {{F_s},2} \right)} &  \cdots  & {Y\left( {{F_s},{T_s}} \right)}  \\
\end{array}} \right]
\end{equation}

Each band is processed by a different 2D psychoacoustic filter. For the implementation of temporal integration (TI), the centre parameter should be different between speech and non-speech frames. The optimal TI parameter, $\alpha_{TI}$, is obtained empirically and is shown in Table \ref{tab:TIpara}.
\begin{table}[!ht]
\renewcommand{\arraystretch}{1.3}
\caption{Temporal Integration Parameter} \label{tab:TIpara}
\centering 
    \begin{tabular}{|c|cc|}
    \hline
     & Speech & Non-speech\\
    \hline
    Low Band  & 4 & 3\\
    High Band & 3 & 2\\
    \hline
    \end{tabular}
\end{table}

\begin{figure}[!h]
\centering
\includegraphics[width=0.7\textwidth]{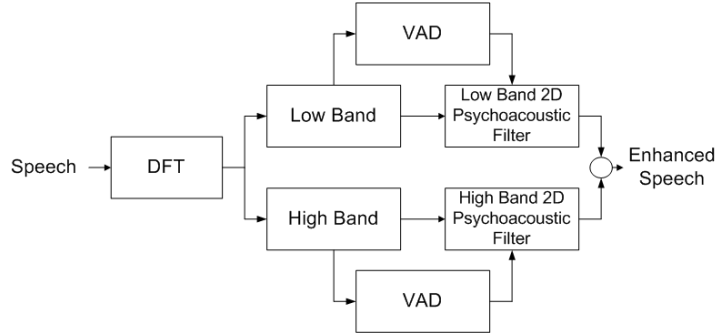}
\vspace{-20pt}
\caption{Block diagram of adaptive 2D psychoacoustic filtering.} \label{fig:adaptive2Dsys}
\end{figure}

\begin{figure}[!h]
\centering
\includegraphics[width=0.49\textwidth]{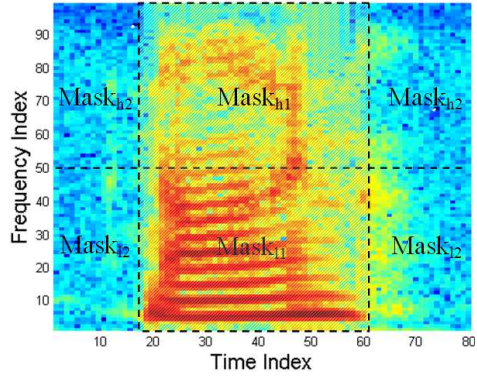}
\caption{Adaptive 2D Psychoacoustic Filtering.}
\label{fig:domainDiv}
\end{figure}

Figure \ref{fig:adaptive2Dsys} illustrates the proposed algorithm. After DFT, the speech spectrogram is equally divided into high and low bands (see Figures \ref{fig:adaptive2Dsys} and \ref{fig:domainDiv}). A voice activity detector (energy ratio test \cite{cohen03noise}) is utilized to distinguish speech/non-speech frames.

\begin{equation}\label{eq:oaef}
\begin{array}{l}
 {\bf Mask }_{OAE}
  = \left[ {\begin{array}{*{20}{c}}
   {{{\bf{0}}_{({F_1} - {F_2}) \times T}}} & {{{\bf{0}}_{ ({F_2} - {F_1}) \times T+1}}}  \\
   {{{\bf{0}}_{({F_1} + {F_2} + 1) \times T}}} & {\begin{array}{*{20}{c}}
   {  \alpha \left( {{F_2},0} \right)} & {} & {} & {  \alpha \left( {{F_2}, - T_{fm}} \right)}  \\
    \vdots  & {} &  {\mathinner{\mkern2mu\raise1pt\hbox{.}\mkern2mu
 \raise4pt\hbox{.}\mkern2mu\raise7pt\hbox{.}\mkern1mu}}  & {}  \\
   {  \alpha \left( {1,0} \right)} & {  \alpha \left( {1, - 1} \right)} & {} & {}  \\
   {1 } & {  \alpha \left( {0, - 1} \right)} &  \cdots  & {  \alpha \left( {0, - T_{fm}} \right)}  \\
   {  \alpha \left( { - 1,0} \right)} & {  \alpha \left( { - 1, - 1} \right)} & {} & {}  \\
    \vdots  & {} &  \ddots  & {}  \\
   {  \alpha \left( { - {F_1},0} \right)} & {} & {} & {  \alpha \left( { - {F_1}, - T_{fm}} \right)}  \\
\end{array}}  \\
\end{array}} \right] \\
 \end{array}
\end{equation}

For each band, two different temporal integration parameters are used. Therefore, there are four different 2D psychoacoustic filters overall in our implementation. As shown in Figure \ref{fig:domainDiv}, four different maskers are adopted for different situations.

In our present implementation, noise is estimated using a
minimum-controlled recursive moving-average noise tracker similar to
the one described in \cite{cohen03noise,MS_joint}. Generally, a
decision on whether a frame contains speech or noise is made based
on the energy ratio test \cite{cohen03noise},
\begin{equation}\label{eq:a2D4}
\frac{{\left| {{P_y}\left( {f_i,t_i} \right)} \right|_t^2}}{{\left| {{P_n}\left( {f_i,t_i} \right)} \right|_{\min }^2}} > \nu
\end{equation}
where $\nu$ is the threshold, $|{{P_n}\left( {f,t} \right)}|_{\min}^2$ is the smoothed minimum noise power within a sliding window which can be tracked efficiently and
$| {{P_y}\left( {f_i,t_i} \right)}|_t^2$ is the smoothed (using adjacent channels) power of the noisy speech \cite{MS_joint}.


Table \ref{tab:atfw2Dlow} (Appendix \ref{ap:2dfilter}) gives the low-band adaptive 2D psychoacoustic filter (without normalization). Here, $\alpha_{TI}^{low}$
is defined as
\begin{equation}\label{eq:a2Dlow}
\alpha _{TI}^{low} = \left\{ {\begin{array}{*{20}{c}}
   4 & {{\rm{Speech}}}  \\
   3 & {{\rm{Non-speech}}}  \\
\end{array}} \right.
\end{equation}

The high-band 2D psychoacoustic filter is given in Table \ref{tab:tfw2Dhigh} in Appendix \ref{ap:2dfilter}. Here, $\alpha_{TI}^{high}$ is defined as
\begin{equation}\label{eq:a2Dhigh}
\alpha _{TI}^{high} = \left\{ {\begin{array}{*{20}{c}}
   3 & {{\rm{Speech}}}  \\
   2 & {{\rm{Non-speech}}}  \\
\end{array}} \right.
\end{equation}

\subsection{Otoacoustic emissions (OAEs)}\label{ssc:oto}
Otoacoustic emissions (OAEs) are clinically important because they are the basis of a simple, non-invasive, test for hearing defects in newborn babies and in children who are too young to cooperate in conventional hearing tests \cite{JLO:8108439,Veros2014}.
OAEs are considered to be related to the amplification function of the cochlea \cite{oto1} and are generated within the inner ear, specifically by the motion of the nerve cells on the basilar membrane within the cochlea as they energetically respond to auditory stimulation \cite{kemp02}. Masking effects can also partially be described by the inner ear, and we assume that OAEs can likewise be calculated using similar equations as masking effects. Previous theoretical studies have suggested that OAEs arise primarily from a linear process of coherent reflection \cite{Zweig1995,Talmadge2000}, which means it can be treated as the `reverberation' of  the input acoustic signal. By using appropriate microphones, we can effectually capture sounds generated by the inner ear itself. Besides, since OAEs are generated by the inner ear, it is logical to assume that the sound (OAEs) can also be captured by the human auditory system, which means the sound we hear is the combination of the original acoustic signal and the OAEs. It has to be noted that the above mentioned phenomena do not necessarily mean that we can acutally hear the OAEs. What we perceive is the result of a series of complicated neuralogical and psychological phenomena. OAEs together with many other psychoacoustic effects (e.g. masking effects, critial bands, etc) help to change the spectrum (or statistics) of the speech, which help to enhace or suppress certain regions of the original speech.



The objective of the proposed algorithm is to recognize speech based on the `actual' speech that is changed to neural spikes by the human auditory system. With OAEs, the new version of speech with OAEs can be modeled as
\begin{equation}\label{eq:oae1}
\tilde {Y}(f,t)=Y(f,t) + M_{OAE}.
\end {equation}
where $M_{OAE}$ represents the amount of OAEs. In our present implementation, OAEs are calculated by
\begin{eqnarray}\label{eq:oae2}
  &&\ M_{OAE} \nonumber  \\
  & = &  \mu \times M_{total} \\
  & = & \mu {\sum_{\Delta t=-T_{bm}}^{T_{fm}}\sum_{\Delta f=-F_1}^{F_2}\alpha(\Delta f,\Delta t)Y(f+\Delta f,t+\Delta t)}.  \nonumber
\end{eqnarray}

The final version of the `new' speech can be calculated by the joint effect of psychoacoustics and OAEs. For a acoustic signal that we hear ($Y(f,t)$), it firstly goes through OAEs, leading to
\begin{eqnarray}\label{eq:oae3}
  &&\ \tilde{Y}_{OAE}(f,t) \nonumber \\
  &=& Y(f,t) + M_{OAE} \\
  &=& Y(f,t) \otimes {\bf Mask}_{OAE}\nonumber.
\end{eqnarray}
where ${\bf Mask}_{OAE}$ is given in Equation \eqref{eq:oaef}. Then, $Y_{OAE}(f,t)$ is further processed by masking effects,
\begin{eqnarray}\label{eq:oae4}
  &&\ \tilde{Y}_{OAE}(f,t) \nonumber \\
  &=& Y_{OAE}(f,t) - M_{psy} \\
  &=& Y_{OAE}(f,t) \otimes {\bf Mask}_{OAE}\nonumber \\
  &=& Y(f,t) \otimes {\bf Mask}_{OAE} \otimes {\bf Mask}_{psy}\nonumber.
\end{eqnarray}

The OAE and psychoacoustic filters are implemented in sequentially in Equation \eqref{eq:oae4}) since OAEs are generated mostly by the inner ear, while psychoacoustic (masking) effects arise mostly from the limits of the auditory nerves immediately proximal. That is, OAEs are first added to the original speech before the mixed speech goes through the entire auditory system.

\section{Theoretical Analysis}\label{sc:2D_theoretical}
\subsection{Complex Spectral Processing}\label{ssc:2D csp}
After being cut into frames and processed by Discrete Fourier Transform (DFT), the speech signal is transformed into the time-frequency domain,
\begin{equation}\label{eq:ta1}
Y\left( {f,t} \right) = {Y_r}\left( {f,t} \right) + i \times {Y_{im}}\left( {f,t} \right)
\end{equation}
where $i$ is the imaginary unit.

Often, only the power or magnitude spectra are extracting from speech in practical applications, and the phase information is simply ignored. However, the phase can encapsulate useful information in speech \cite{phase1asr,phase2asr}. Our proposed algorithm works directly in the time-frequency domain, including phase, in the noise removing process.
\begin{equation}\label{eq:ta3}
\begin{array}{l}
 \tilde Y\left( {f,t} \right) = Y\left( {f,t} \right) \ast Mask \\
  = \left[ {{Y_r}\left( {f,t} \right) + i \times {Y_{im}}\left( {f,t} \right)} \right] \ast Mask \\
  = {Y_r}\left( {f,t} \right) \ast Mask + i \times {Y_{im}}\left( {f,t} \right) \ast Mask \\
 \end{array}
\end{equation}
where $\ast$ is the convolution operator.

\subsection{Double Transform}\label{ssc:double}
Typically, each frame of speech in time is transformed into the frequency domain using the discrete Fourier transform (DFT). One key difference between our 2D psychoacoustic filters and normal spectral filtering is that 2D psychoacoustic filters are implemented by convolution in the time-frequency domain. Therefore, the analysis of high-pass or low-pass filters should be made in terms of the 2D frequency spectrum of the time-frequency domain speech signal. The 2D Fourier transform of the time-frequency domain speech signal is denoted as a double transform in later discussion. While the high-pass 2D psychoacoustic filter preserves high-frequency signals, it also attenuates signals in terms of the double transform spectrum, i.e., the 2D Fourier transform of the time-frequency domain signal ($Y(f,t)$).

\begin{figure}[!h]
    \setlength{\abovecaptionskip}{0pt}
    \flushleft
    \centering
    \subfigure[Babble]      {\label{fig:2dft:a} 
    \includegraphics[width=0.4\textwidth]{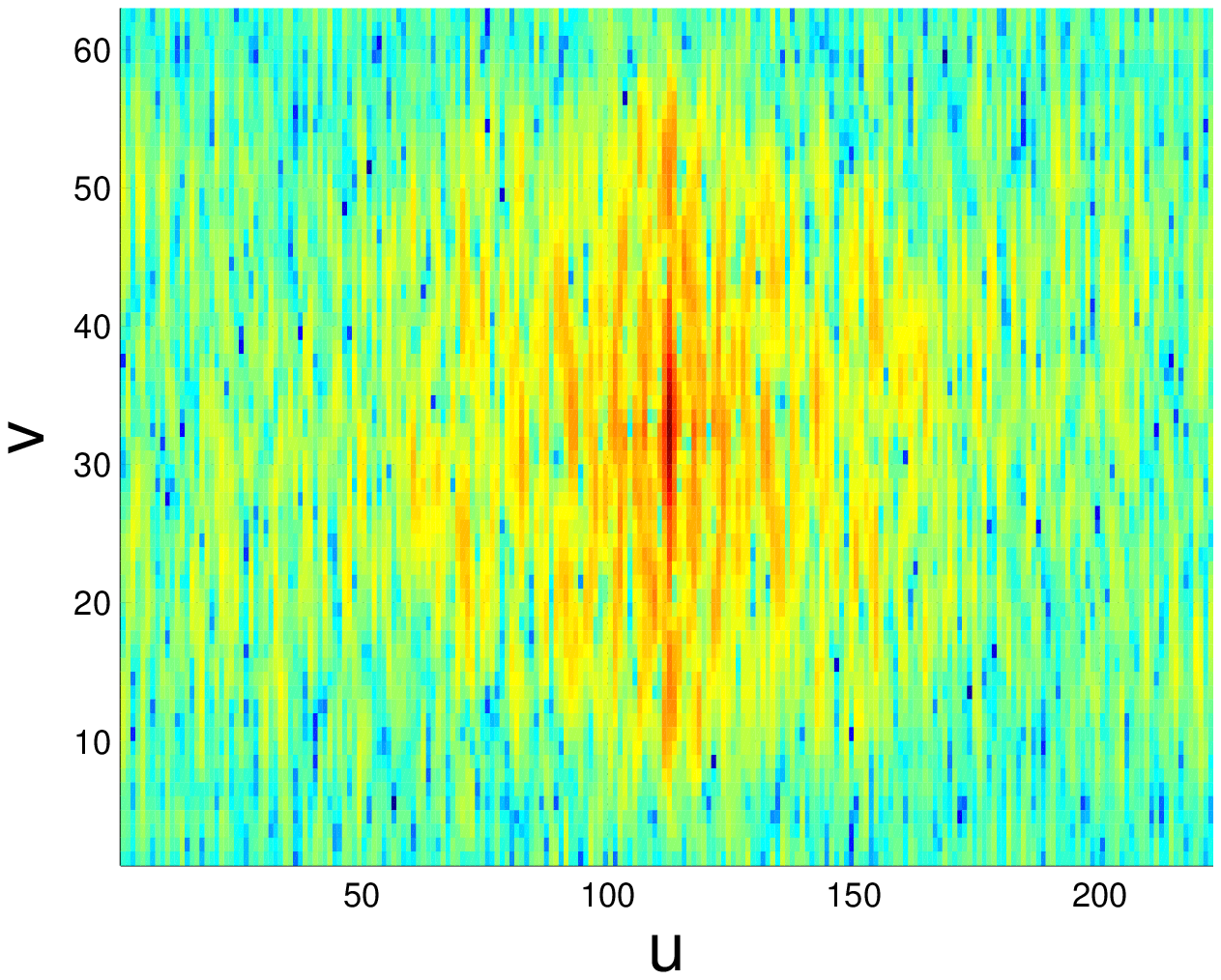}}
    \subfigure[Restaurant]  {\label{fig:2dft:b} 
    \includegraphics[width=0.4\textwidth]{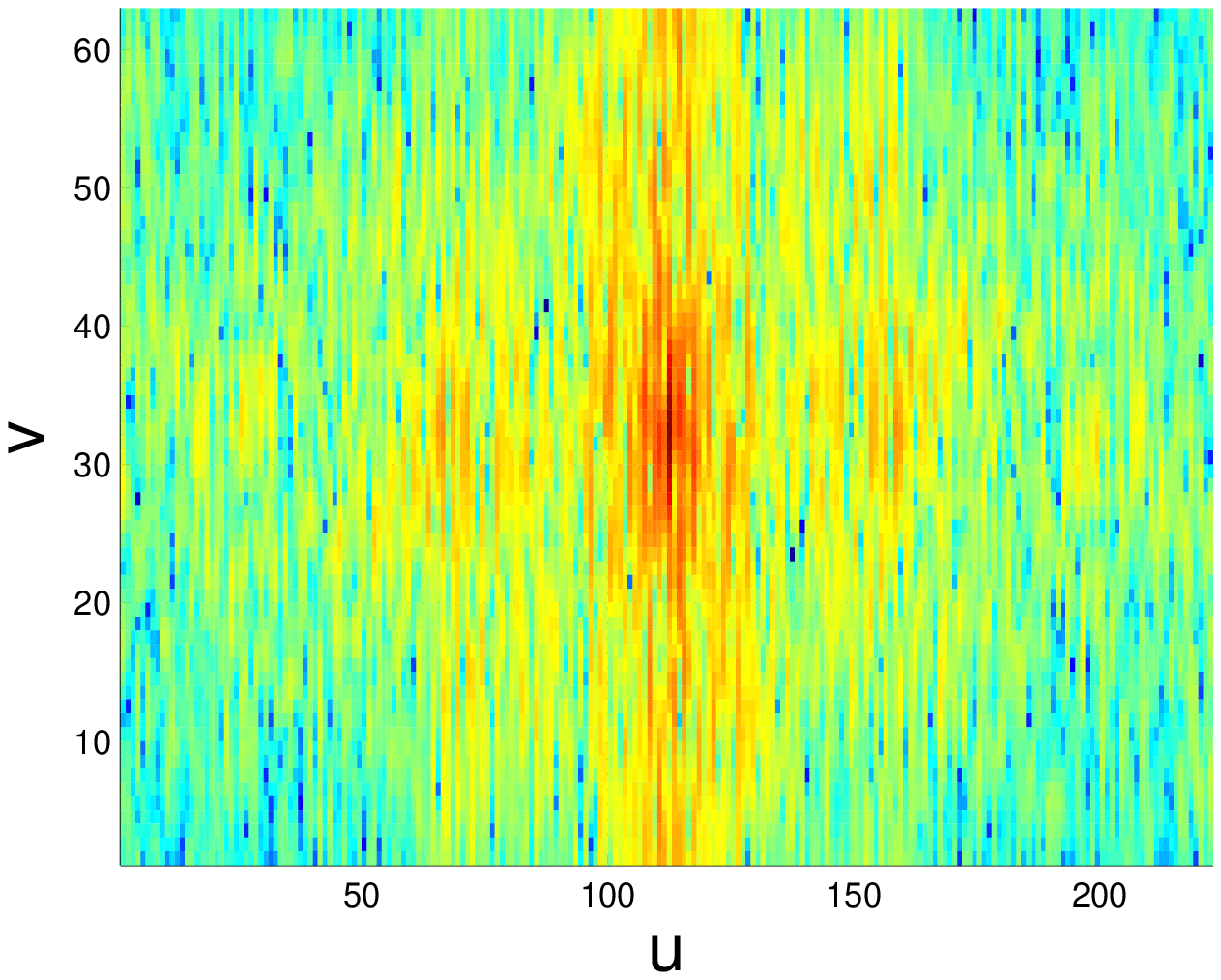}}
    \subfigure[Clean Speech]      {\label{fig:2dft:c} 
    \includegraphics[width=0.4\textwidth]{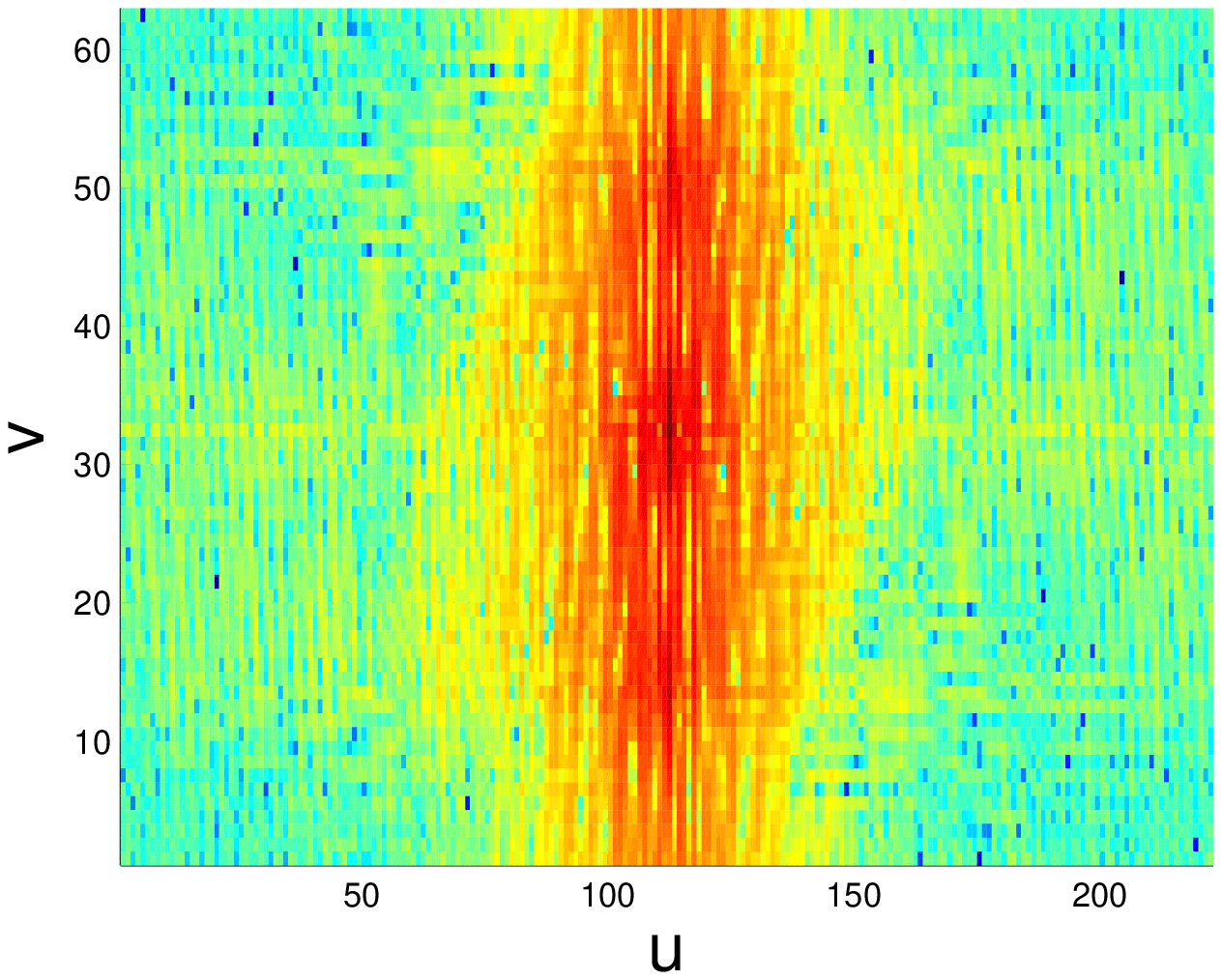}}
    \subfigure[2D psychoacoustic filter \cite{Dai12sc}]  {\label{fig:2dft:d} 
    \includegraphics[width=0.4\textwidth]{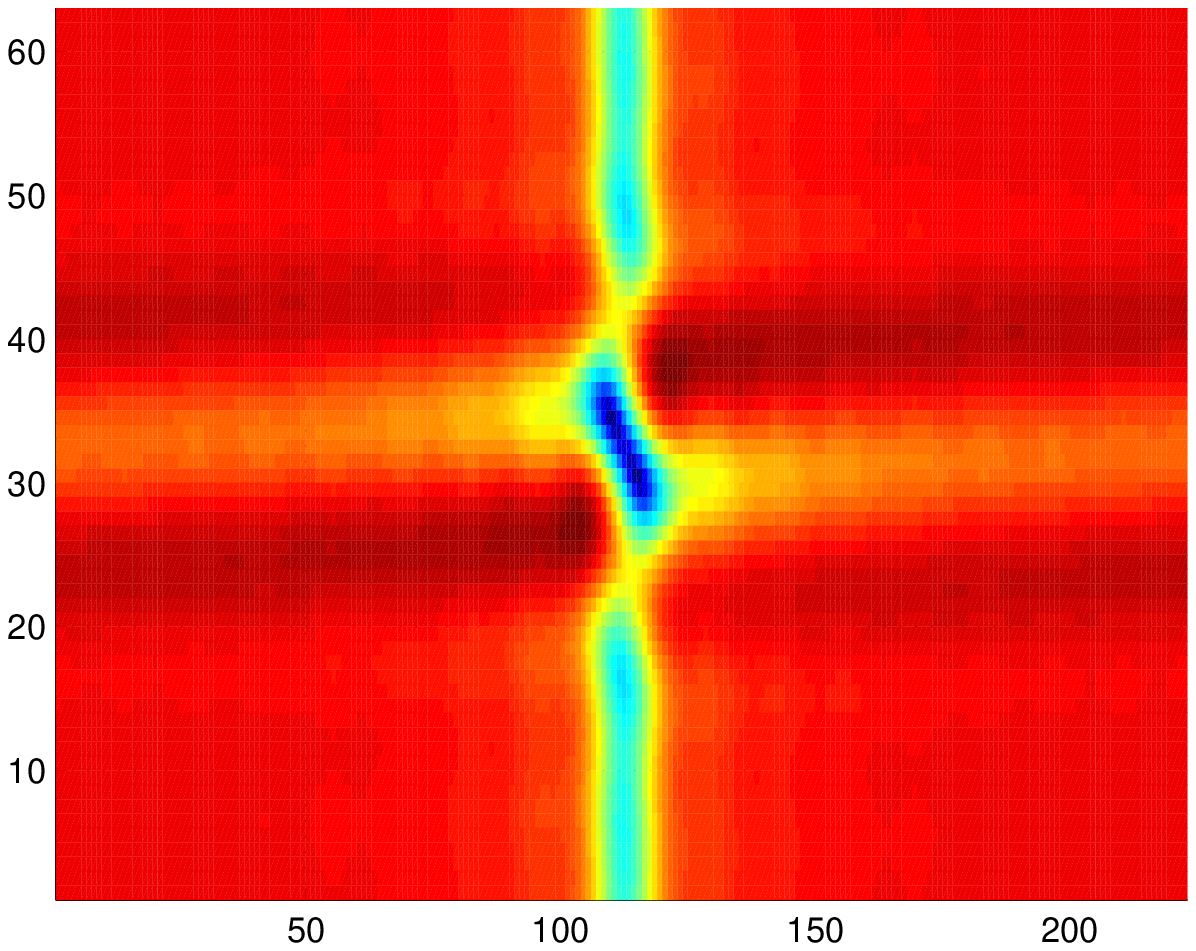}}
    \caption{Double transform spectrum, $u$ and $v$ are the 2D spatial frequencies.}
    \label{fig:cleanF} 
\end{figure}

Figures \ref{fig:2dft:a} and \ref{fig:2dft:b} shows the double transform spectrum of two different kinds of noise: babble and restaurant (taken from the AURORA2 databse). We provide the double transform spectrum of clean speech and the frequency reponse of the 2D psychoacoustic filter (introduced in our previous paper \cite{Dai12sc}) in Figure \ref{fig:cleanF}. Speech and noise behave very differently in the double transform domain, where speech is more concentrated in the centre column. Based on the double transform spectrum, we can analyze qualitatively which type of noise to which our proposed algorithm is most suited. Double transform analysis allows us to explain why our empirical results are better for certain noise types since the adaptive psychoacoustic filter proposed in this paper adopts different parameters for different frequency bands. Detailed analysis using speech recognition results is given in Section \ref{ssc:double2}.

\section{Results and Discussion}\label{sc:2D_result}
\subsection{Data and Methods}
\subsubsection{System Description}\label{ssc:aurora2}
Evaluation is carried out using the AURORA2 database \cite{Pearse2000}. The AURORA2 data are based on a version of the original TIDigits (available from LDC) downsampled to 8 kHz \cite{Pearse2000, Leonard1984}. The database provides two different training patterns, i.e., a clean training condition and a multi-training condition. The clean training set has no noise added and consists of 8440 utterances recorded from 55 male and 55 female adults. In total, 4004 utterances from 52 male and 52 female speakers are split equally into 4 subsets, with all speakers present in each subset. In the multi-training condition (i.e., `multi-condition' training) set, four types of noise are added at SNR levels 20 dB, 15 dB, 10 dB, 5 dB, 0 dB, and -5 dB. The database covers eight different noise types, i.e. subway, babble, car, exhibition, restaurant, street, airport and train station (provided in test set A and B). Additionally, the database provides a telephone speech test set. In test set C, two types of noise (subway and street) processed by the modified intermediate reference system (MIRS) filter are added, which simulates the frequency characteristics of a telecommunication terminal \cite{Pearse2000, Leonard1984}.

 The same recognizer is used for both the proposed algorithm and the comparison targets. Each digit is modeled by a simple left-to-right 18-state HMM model (including two non-emitting states), with 3 Gaussian mixtures per state. Two pause models are defined. One is ``sil", which has 3 HMM states and models the pauses before and after each utterance, the other is ``sp", which is a single state model (tied with the middle state of ``sil") and models pauses among words \cite{Pearse2000,Dai12sc}.

\begin{figure}[!h]
    \centering
    \includegraphics[width=0.8\textwidth]{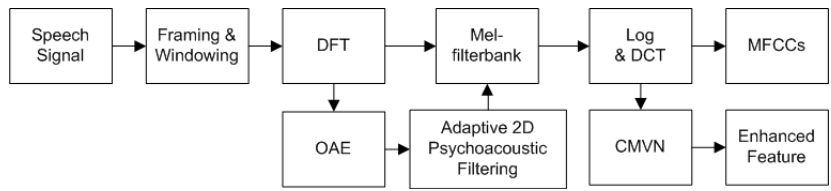}
    \vspace{-25 pt}
    \caption{System Diagram of the proposed algorithm}
    \label{fig:algsys}
\end{figure}

Our proposed algorithm is developed based on Mel-frequency cepstral coefficients (MFCCs). The scripts provided in the AURORA2 database are used for training and testing. The same recognizer is used for both the proposed algorithm and the comparison targets. Specifically, each digit is modeled by a simple left-to-right 18-state (including two non-emitting states) hidden Markov model, with 3 Gaussian mixtures per state. Two pause models are defined: {\em sil} has 3 HMM states and models the pauses before and after each utterance, and {\em sp} has a single state tied with the middle state of {\em sil} and models pauses among words \cite{Pearse2000,Dai12sc}. The baseline results are based on the standard 13 MFCCs together with the corresponding velocity and acceleration parameters, denoted as MFCC(39). Figure \ref{fig:algsys} gives the diagram of the proposed algorithm.

Evaluation is performed in terms of recognition rate. Experimental results are averaged over 0 dB - 20 dB, denoted as Avg 0-20. Relative improvement is defined as
\begin{equation}\label{eq:rel}
R_{im}=\frac {r_{p}-r_{t}}{r_{t}} \times 100\%
\end {equation}
where $R_{im}$ is the relative improvement; $r_{p}$ is the recognition rate of our proposed algorithm; $r_{t}$ is the recognition rate of the comparition target.


\subsubsection{Comparison Targets}\label{ssc:comt}
Three sets of comparisons are presented to show the effectiveness of our proposed algorithm. First, we compare our proposed algorithm with earlier implementations of psychoacoustic filters. Then we compare our proposed algorithm with MFCC, forward masking, lateral inhibition (LI), and cepstral mean \& variance normalization (CMVN). The final set of comparisons is made against state-of-the-art noise removal methods frequently used in ASR systems namely RelAtive SpecTrAl (RASTA) noise removal \cite{RASTA}, minimum mean square error (MMSE) \cite{MMSE}, mean variance normalization \& ARMA filtering (MVA, where the ARMA filter is an autoregressive moving average filter) \cite{MVA}, and the ETSI Advanced FrontEnd (AFE) \cite{AFE}.


The MMSE estimator was first proposed for speech enhancement in 1984 \cite{MMSE}. The algorithm models speech and noise spectra as statistically independent Gaussian random variables. By minimizing the mean square error, the problem is formulated as
\begin{equation}\label{eq:logmmse0}
    \min {\left[ { \left| Y(f,t) \right| -  | {\tilde X}(f,t) |} \right]^2}.
\end{equation}

The Relative Spectra (RASTA) was proposed by Hermansky in 1994 and is based on the fact that human perception tends to react to the relative value of an input \cite{RASTA}. The transfer function of the RASTA filter is
\begin{equation}\label{eq:rasta}
H(z)=0.1z^4\times\frac{2+z^{-1}-z^{-3}-2z^{-4}}{1-0.98z^{-1}}.
\end {equation}

MVA is a very effective cepstral-domain filtering algorithm. It works by implementing an ARMA cepstral filter (i.e., `lifter') and manages to effectively improve ASR performance empirically \cite{MVA}. The AFE algorithm is an improved form of Wiener filter, which can adapt to the noise to a certain extent \cite{AFE}.

\begin{figure}[!h]
    \setlength{\abovecaptionskip}{0pt}
    \flushleft
    \centering
    \subfigure[]      {\label{fig:double_res:a} 
    \includegraphics[width=0.4\textwidth]{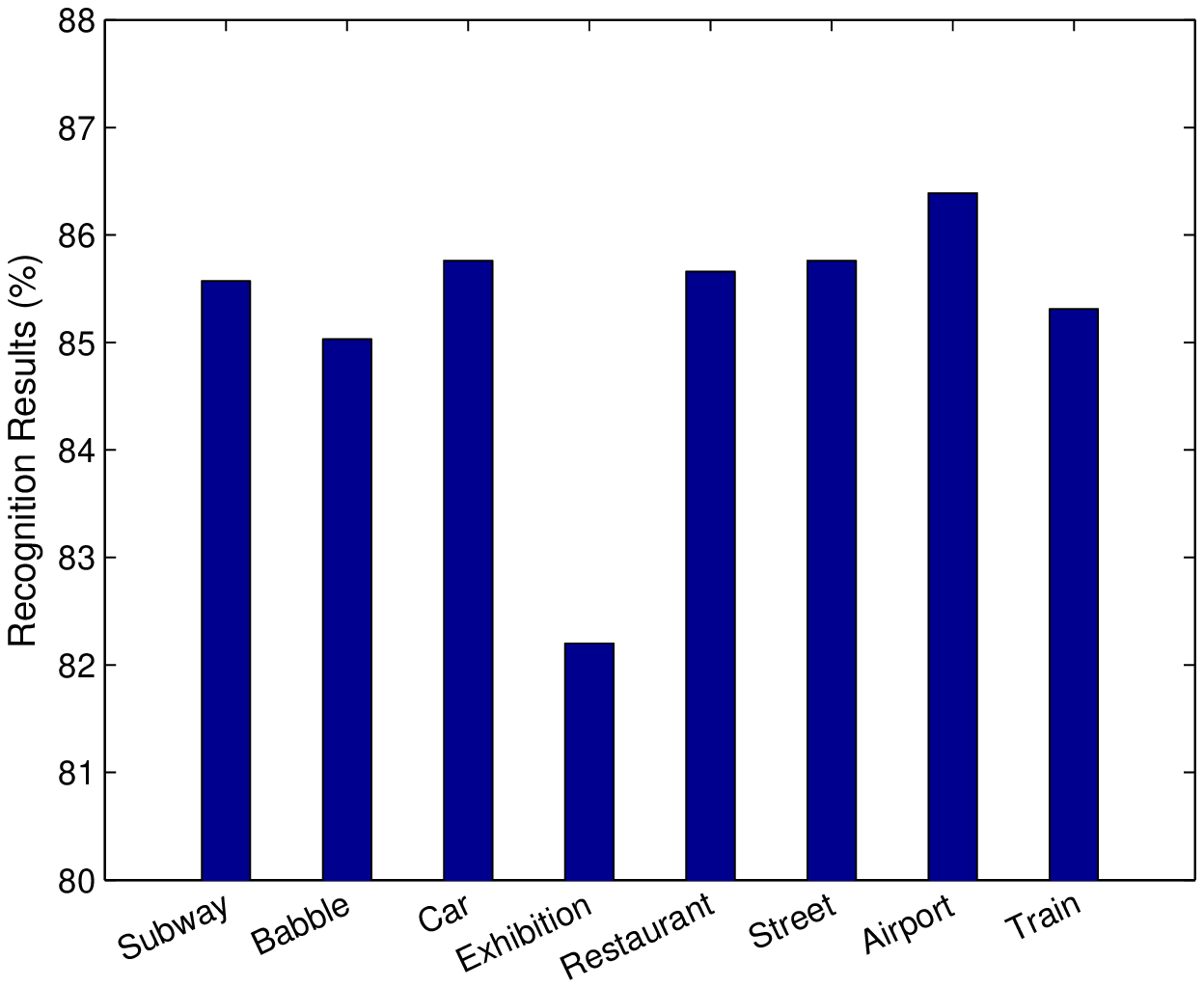}}
    \subfigure[]  {\label{fig:double_res:b} 
    \includegraphics[width=0.4\textwidth]{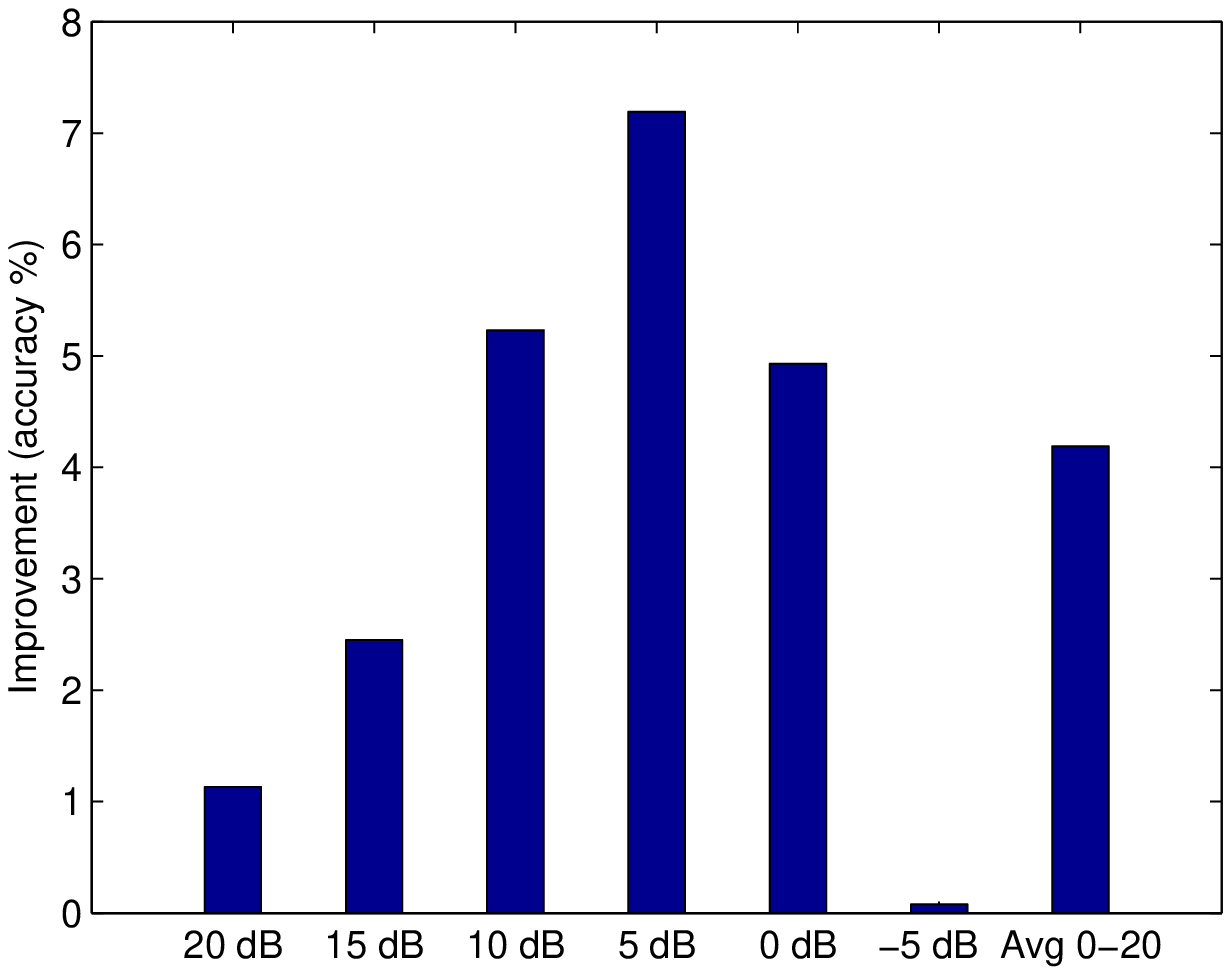}}
    \caption{Recognition results of TFW 2D psychoacoutic filter (\%) for the clean training condition: (a) The recognition results of different noise types; (b) The improvement of the recognition result for Airport noise over Exhibition noise.}
    \label{fig:double_res} 
\end{figure}

\subsection{Double Transform}\label{ssc:double2}
In Section \ref{ssc:double}, we proposed a novel double transform analysis technique, which can be used to quanlitatively analyze the ASR performance of psychoacoustic filters in terms of the property of the proposed filters, e.g. high pass or low pass. For the proposed adaptive 2D psychoacoustic filter, the final recognition accuracy is a result of the joint effect of both bands, which would be very difficult to analyze otherwise. Therefore, we take the temporal frequency warped 2D psychoacoustic filter \cite{Dai12sc} as example to show the general steps of double transform analysis. Table \ref{tab:c7tfw2D} gives the ASR experimental results based on the AURORA2 database.


\begin{table*}[!h]
\caption{Recognition results of TFW 2D psychoacoutic filter (\%) for the clean training condition. } \label{tab:c7tfw2D} \centering \vspace{5pt}
    \begin{tabular}{|c|c||ccccccc||c|}
    \hline
     & Noise Type & Clean & 20 & 15 & 10 & 5 & 0 & -5 & Avg 0-20\\
    \hline
    \cline{1-10}            	&   Subway      	&   99.45   &   97.45   &   95.58   &   92.29   &   81.79   &   60.73   &   27.23   &   85.57  \\
    \cline{2-10}    Set A   	&   Babble      	&   99.21   &   98.16   &   96.61   &   93.47   &   81.95   &   54.96   &   23.64   &   85.03  \\
    \cline{2-10}            	&   Car         		&   99.34   &   98.06   &   96.51   &   92.63   &   82.52   &   59.08   &   22.93   &   85.76  \\
    \cline{2-10}            	&   Exhibition  	&   99.63   &   97.35   &   94.72   &   88.92   &   75.93   &   54.09   &   25.33   &   82.20  \\
    \hline
    \cline{1-10}            	&   Restaurant  	&   99.45   &   98.59   &   97.02   &   93.06   &   81.92   &   57.69   &   28.74   &   85.66  \\
    \cline{2-10}    Set B   	&   Street          	&   99.21   &   97.88   &   96.13   &   92.17   &   82.38   &   60.25   &   26.57   &   85.76  \\
    \cline{2-10}           	&   Airport     	&   99.34   &   98.48   &   97.17   &   94.15   &   83.12   &   59.02   &   25.41   &   86.39  \\
    \cline{2-10}           	&   Train       		&   99.63   &   98.06   &   96.54   &   92.75   &   82.78   &   56.4    &   23.51   &   85.31  \\
    \hline
    \cline{1-10}    Set C   	&   Restaurant  	&   99.36   &   97.30   &   94.90   &   89.90   &   77.34   &   51.55   &   21.25   &   82.20  \\
    \cline{2-10}            	&   Street      	&   99.27   &   97.28   &   95.59   &   90.02   &   78.96   &   54.90   &   23.31   &   83.35  \\
    \hline
    \cline{1-10} \multicolumn{2}{|c||}{Avg} &   99.38   &   97.77   &   95.94   &   91.61   &   80.42   &   56.26   &   24.37   &   84.40  \\
    \hline
    \end{tabular}
\end{table*}

Clearly, the TFW 2D filter is best fit for airport noise. It posesses a peak at the centre column, which can be blocked by the 2D psychoacoustic filter (see Figure \ref{fig:2dft:a}) and obtains 86.39\%, also shown in Figure \ref{fig:double_res}. The double transform spectrum of exhibition noise covers a large amount of the centre column and appears very similar to speech. Contrariwise, the recognition results given exhibition noise is worse than other noise types at 82.60\%. Figure \ref{fig:double_res:b} shows the ASR performance difference in terms of recognition rate (Airport noise condition ASR result minus the corresponding Exhibition noise condition result). It can be seen that the 2D psychoacoustic filter yield consistently better result for all the given SNR levels in Airport noise condition. In particular, more improvements are obtained at SNRs from 10 dB $\sim$ 0 dB. This is mainly due to the fact that ASR system yields nearly perfect performance ($>90\%$) at high SNR levels (e.g. $SNR>10 dB$), which leaves little place for improvement. For extremely low SNR levels, noise becomes dominant, which possesses stronger energy than speech. Thus, ASR systems obtain terrible performance at this condition.



\subsection{Experimental Results}\label{ssc:in2D_results}

\begin{figure}[!h]
    \centering
    \includegraphics[width=0.48\textwidth]{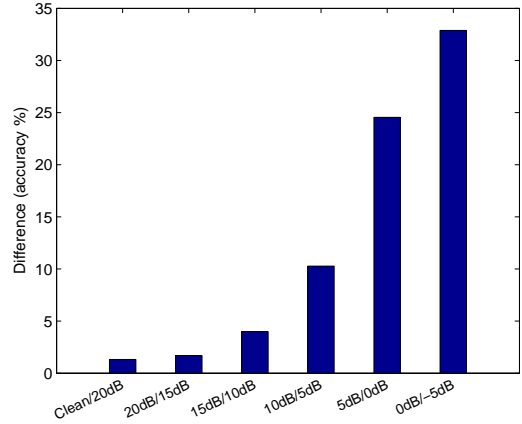}
    \caption{The ASR performance difference between neighboring SNR levels.}
    \label{fig:adapt_res}
\end{figure}

Detailed experimental results for the proposed adaptive 2D psychoacoustic filter are given in Tables \ref{tab:c4tfw2Dclean} and \ref{tab:c4tfw2Dmulti} including the results for different noise types and SNR levels. The AURORA2 database provides 7 different SNR levels. As SNR drops, the recognition rate degrades at increasing speed. Figure \ref{fig:adapt_res} gives the recognition rate 'drop' between neighboring SNR levels, e.g. Clean Vs. 20 dB (denoted as Clean/20dB). It can be seen that at high SNR levels, e.g. $SNR>10 dB$, the addition of noise causes relatively less degrade to the system performance. However, as SNR drops below 10 dB, the performance of the ASR system significantly drops, $14\%\sim30\%$.

\begin{table*}[h]
\caption{Recognition Results of Proposed Algorithm for Clean Training Condition (\%)} \label{tab:c4tfw2Dclean} \centering \vspace{5pt}
    \begin{tabular}{|c|c||ccccccc||c|}
    \hline
     & Noise Type & Clean & 20 & 15 & 10 & 5 & 0 & -5 & Avg 0-20\\
    \hline
    \cline{1-10}        &   Subway      &   99.42 &  97.73 &  95.98 &  92.60 &  84.03 &  64.26 &  29.66 &  86.92  \\
    \cline{2-10}Set A   &   Babble      &   99.15 &  98.40 &  96.98 &  93.92 &  82.47 &  54.26 &  21.98 &  85.21  \\
    \cline{2-10}        &   Car         &   99.28 &  98.24 &  96.87 &  92.66 &  83.42 &  57.62 &  21.74 &  85.76  \\
    \cline{2-10}        &   Exhibition  &   99.72 &  97.72 &  94.63 &  89.20 &  76.55 &  54.92 &  28.48 &  82.60  \\
    \hline
    \cline{1-10}        &   Restaurant  &   99.42 &  98.68 &  97.27 &  93.43 &  83.21 &  58.09 &  27.11 &  85.86  \\
    \cline{2-10}Set B   &   Street      &   99.15 &  97.76 &  96.34 &  92.53 &  83.04 &  59.64 &  25.85 &  86.14  \\
    \cline{2-10}        &   Airport     &   99.28 &  98.39 &  97.20 &  94.48 &  83.63 &  59.44 &  24.19 &  86.63  \\
    \cline{2-10}        &   Train       &   99.72 &  98.18 &  96.64 &  93.00 &  83.37 &  56.53 &  22.03 &  85.54  \\
    \hline
    \cline{1-10}Set C   &   Restaurant  &   99.36 &  97.85 &  95.70 &  90.76 &  81.24 &  56.03 &  23.46 &  84.32  \\
    \cline{2-10}        &   Street      &   99.15 &  97.58 &  95.92 &  90.99 &  79.96 &  54.69 &  22.19 &  83.83  \\
    \hline
    \cline{1-10} \multicolumn{2}{|c||}{Avg} & 99.37 & 98.05 & 96.35 &  92.36 &  82.09 &  57.55 &  24.67 & 85.28 \\
    \hline
    \end{tabular}
\end{table*}

\begin{table*}[!h]
\caption{Recognition Results of Proposed Algorithm for Multi Training Condition (\%)} \label{tab:c4tfw2Dmulti} \centering \vspace{5pt}
    \begin{tabular}{|c|c||ccccccc||c|}
    \hline
     & Noise Type & Clean & 20 & 15 & 10 & 5 & 0 & -5 & Avg 0-20\\
    \hline
    \cline{1-10}        	&   Subway	        &   98.83  & 98.25 &  97.64 &  96.41 &  93.58 &  81.64 &  54.44 &  93.50  \\
    \cline{2-10}Set A       &   Babble          &   98.88  & 98.49 &  97.97 &  97.04 &  91.90 &  74.03 &  39.90 &  91.89  \\
    \cline{2-10}        	&   Car         	&   98.75  & 98.21 &  97.52 &  96.42 &  91.68 &  77.01  & 42.20  & 92.17  \\
    \cline{2-10}        	&   Exhibition      &   99.14 &  98.52 &  97.59 &  94.66 &  88.28 &  73.99 &  48.90 &  90.61  \\
    \hline
    \cline{1-10}        	&   Restaurant  	&   98.83 &  98.56 &  98.04 &  97.14 &  91.93 &  76.54 &  44.03 &  92.44  \\
    \cline{2-10}Set B       &   Street         	&   98.88 &  98.46 &  97.79 &  95.77 &  90.51 &  76.36 &  45.47  & 91.78  \\
    \cline{2-10}        	&   Airport        	&   98.75 &  98.42 &  97.97 &  97.02 &  92.48 &  78.14 &  43.81  & 92.81  \\
    \cline{2-10}        	&   Train           &   99.14 &  98.86 &  97.99 &  96.79 &  91.61 &  75.04 &  41.31  & 92.06  \\
    \hline
    \cline{1-10}Set C       &   Restaurant  	&   98.77  & 98.16 &  97.54  & 96.38 &  92.05  & 78.42  & 46.45  & 92.51  \\
    \cline{2-10}        	&   Street      	&   98.85  & 98.28 &  97.70  & 95.56  & 90.05  & 75.15 &  40.72  & 91.35  \\
    \hline
    \cline{1-10} \multicolumn{2}{|c||}{Avg}     & 98.88 & 98.42 & 97.78 & 96.32 & 91.41 & 76.63 & 44.72 & 92.11 \\
    \hline
    \end{tabular}
\end{table*}

Experimental results for coparison targets are given in Tables \ref{tab:CompTargets1} and \ref{tab:CompTargets2}.  All comparison methods are implemented with MFCC(39). Experimental results are averaged over SNR of 0 dB to 20 dB denoted as Avg 0-20. `Rel. Imp.' stands for relative improvements in terms of recognition rate (see Equation \eqref{eq:rel}).
\begin{table*}[htb]
\renewcommand{\arraystretch}{1.3}
\caption{Recognition results for comparison targets under clean
training condition (\%)} \label{tab:CompTargets1} \centering
\vspace{5pt}
    \begin{tabular}{|c|ccccccc|c|}
    \hline
     SNR/dB & Clean & 20 & 15 & 10 & 5 & 0 & -5 & Avg 0-20\\
    \hline
    MFCC(39)    &  99.36 &  97.37 &  93.51 &  81.16 &  56.02 &  28.39 &  13.04 &  71.29\\
    FM          &  99.03 &  97.02 &  93.91 &  85.89 &  68.24 &  41.65 &  21.30 &  77.34\\
    LI          &  99.42 &  97.19 &  94.23 &  83.29 &  60.92 &  34.21 &  17.07 &  73.97\\
    CMVN        &  99.32 &  96.97 &  94.32 &  87.59 &  71.20 &  38.84 &  13.90 &  77.78\\
    \hline
    TW-2D       &  99.33 &  97.47 &  95.59 &  90.22 &  75.70 &  42.85 &  14.41 &  80.36\\
    TFW-2D      &  99.38 &  97.77 &  95.94 &  91.61 &  80.42 &  56.26 &  24.37 &  84.40\\
    \hline
    \end{tabular}
\end{table*}

\begin{table*}[h]
\renewcommand{\arraystretch}{1.3}
\caption{Recognition results for comparison targets under multi training condition (\%)} 
\label{tab:CompTargets2} \centering
\vspace{5pt}
    \begin{tabular}{|c|ccccccc|c|}
    \hline
     SNR/dB     & Clean   & 20 & 15 & 10 & 5 & 0 & -5 & Avg 0-20\\
    \hline
    MFCC(39)    &   99.11 &  98.18 &  97.60 &  95.52 &  87.61 &  60.37 &  26.83 &  87.85\\
    FM          &   98.74 &  98.16 &  97.47 &  95.25 &  87.19 &  59.32 &  25.46 &  87.48\\
    LI          &   99.13 &  98.19 &  97.62 &  95.53 &  88.06 &  61.93 &  26.59 &  88.26\\
    CMVN        &   98.94 &  98.51 &  97.89 &  96.27 &  91.06 &  74.81 &  42.63 &  91.71\\
    \hline
    TW-2D       &   99.05 &  98.57 &  97.90 &  96.30 &  91.08 &  73.41 &  38.57 &  91.45\\
    TFW-2D      &   98.87 &  98.35 &  97.80 &  96.09 &  91.09 &  75.73 &  43.86 &  91.81\\
    \hline
    \end{tabular}
\end{table*}

The relative improvements in terms of Avg 0-20 are given in Tables \ref{tab:c4RelaComp1} and \ref{tab:c4RelaComp2}.
\begin{table}[htb]
\renewcommand{\arraystretch}{1.3}
\caption{Relative Improvements under clean training condition (\%)}
\label{tab:c4RelaComp1} \centering \vspace{5pt}
    \begin{tabular}{|c|c|cc|cc|}
    \hline
     SNR/dB & Clean & Avg 0-20 & Rel. Imp & -5 & Rel. Imp\\
    \hline
    MFCC(39)   & 99.36  & 71.29 &  19.62 &  13.04 &  90.03\\
    FM         & 99.03  & 77.34 &  10.27 &  21.30 &  16.34\\
    LI         & 99.42  & 73.97 &  15.29 &  17.07 &  45.17\\
    CMVN       & 99.32  & 77.78 &  9.64  &  13.90 &  78.27\\
    \hline
    TW-2D      & 99.33  & 80.36 &  6.12  &  14.42 &  71.84\\
    TFW-2D     & 99.38  & 84.40 &  1.04  &  24.37 &  1.68\\
    \hline
    \end{tabular}
\end{table}
\begin{table}[htb]
\renewcommand{\arraystretch}{1.3}
\caption{Relative Improvements under multi training condition (\%)}
\label{tab:c4RelaComp2} \centering \vspace{5pt}
    \begin{tabular}{|c|c|cc|cc|}
    \hline
     SNR/dB & Clean & Avg 0-20 & Rel. Imp & -5 & Rel. Imp\\
    \hline
    MFCC(39)  &  99.11 &  87.85  & 5.22  &  26.83 &  71.93\\
    FM        &  98.74 &  87.48  & 5.67  &  25.46 &  81.19\\
    LI        &  99.13 &  88.26  & 4.73  &  26.59 &  73.49\\
    CMVN      &  98.94 &  91.74  & 0.76  &  42.64 &  8.18\\
    \hline
    TW-2D     &  99.05 &  91.45  & 1.08  &  38.57 &  19.60\\
    TFW-2D    &  98.67 &  91.81  & 0.69  &  43.86 &  5.18\\
    \hline
    \end{tabular}
\end{table}

\subsection{Clean Training Condition}\label{ssc:2D_clean}
\begin{figure}[!h]
    \setlength{\abovecaptionskip}{0pt}
    \flushleft
    \centering
    \subfigure[Clean Training Condition]      {\label{fig:2d_clean} 
    \includegraphics[width=0.4\textwidth]{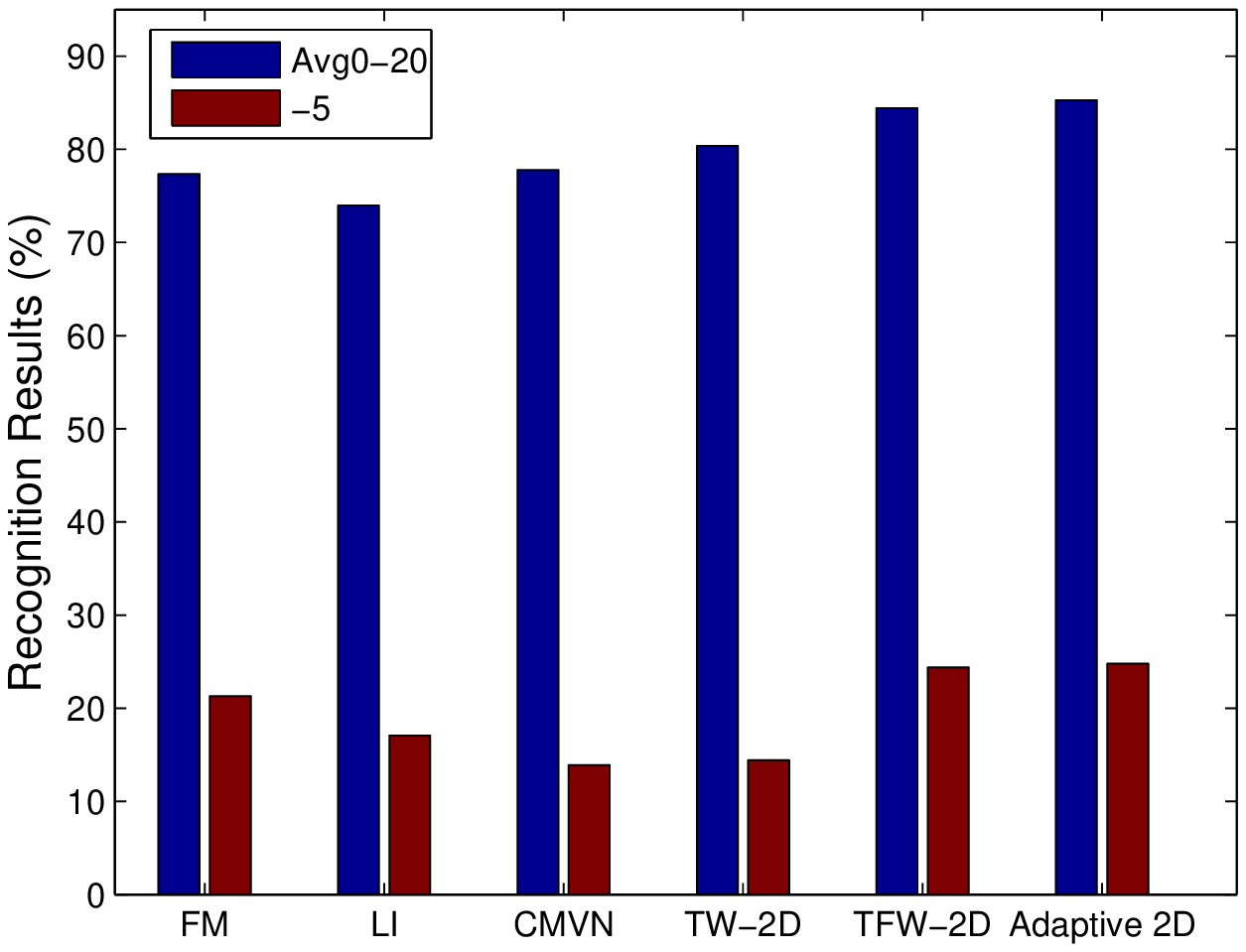}}
    \subfigure[Multi Training Condition]  {\label{fig:2d_multi} 
    \includegraphics[width=0.4\textwidth]{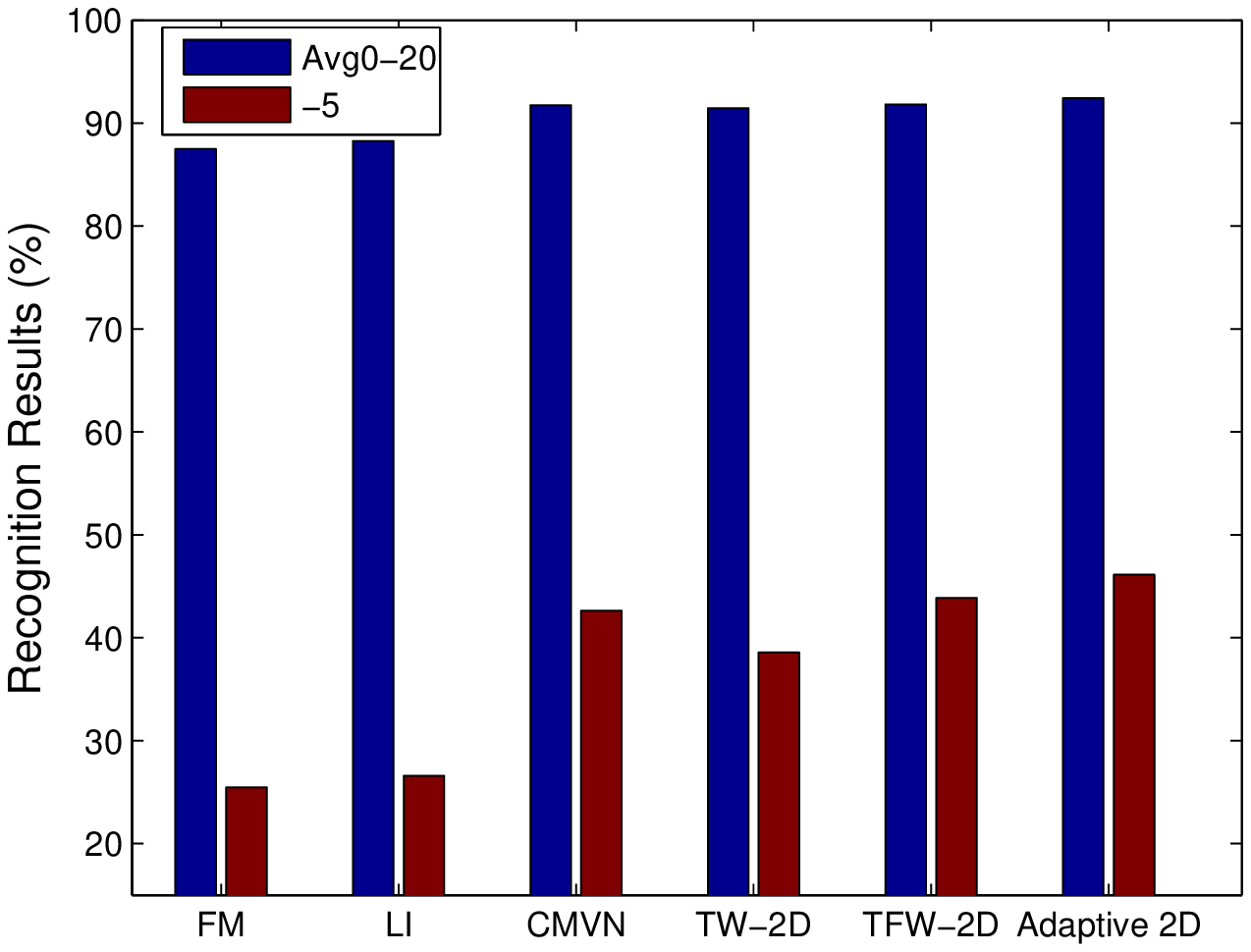}}
    \caption{Experimental results for clean and multi training conditions.}
    \label{fig:double_res} 
\end{figure}

Our proposed algorithm clearly outperforms the other methods , overviewed in Figure \ref{fig:2d_clean}. Compared with MFCC(39), the advantage of the proposed algorithm is obvious. The relative improvement at Avg 0-20 is 19.62\% and at SNR of -5 dB it becomes 90.03\%. For FM, LI and CMVN, the relative improvements at Avg 0-20 are 10.27\%, 15.29\% and 9.64\%. At the SNR -5 dB, the relative improvements are 16.34\%, 45.17\% and 78.27\% respectively.

We propose three different 2D psychoacoustic filters: TW-2D, TFW-2D, and the adaptive 2D psychoacoustic filter. The relative improvements for TW-2D are 6.12\% and 71.84\% for Avg 0-20 and SNR -5 dB respectively. For TFW-2D, the relative improvements are 1.04\% and 1.68\% for Avg 0-20 and SNR of -5 dB respectively.

In order to give a better view of the speech recognition results, we give the statistical test results (Cohen's d) in Table \ref{tab:cohen}. 

It can be seen that our proposed algorithm shows significantly better results. As mentioned earlier, the clean condition
results are very high (around 99\%). Therefore, the difference between the results from different algorithms are relatively small and most of the Cohen's $d$ effect sizes are below 0.5. However, we can see that the clean test result for FM is much worse than others. For Avg 0-20 and -5 dB, the Cohen's $d$ values are mostly larger than 3 (MFCC, FM, LI, and CMVN), which corresponds to $p$-values
at $10^{-4}\sim 10^{-5}$ level. 

\begin{table}[htb]
\renewcommand{\arraystretch}{1.3}
\caption{Statistical test result for comparison targets (Cohen's d).}
\label{tab:cohen} \centering 
    \begin{tabular}{|c|ccc|}
    \hline
            & Clean & Avg 0-20  & -5 \\
    \hline
    MFCC(39)  &  0.3750 & 5.6746  & 3.9431  \\
    FM        &  0.9913  & 5.8136 & 2.9527  \\
    LI        &  0.0100  & 7.2377 & 3.9009  \\
    CMVN      &  0.4713 & 3.5766  & 3.5004  \\
    \hline
    TW-2D     &  0.3904  & 2.1232 & 3.1436 \\
    TFW-2D    &  0.6252  & 1.2192 & 0.2134  \\
    \hline
    \end{tabular}
\end{table}

\subsection{Multi Training Condition}\label{ssc:2D_multi}
There are two training conditions in the AURORA2 database, clean and multi-training conditions. For the multi training condition, since noisy speech is used to train HMMs, the recognition results are all
very good, even achieving about 80\% recognition rate at SNR of 5
dB. The corresponding Cohen's d sizes are all below 1. Therefore large or statistically significant improvements at this level are not very
possible. However, the proposed algorithm still manages to get very
promising results. Figure \ref{fig:2d_multi} shows the relative
improvements of the proposed algorithm over all the comparison
targets.

It can be seen that the proposed algorithm obtains significant improvements. In terms of Avg 0-20, the relative improvements are 5.22\% over MFCC(39), 5.67\% over FM,
4.73\% over LI, 0.76\% over CMVN. For SNR -5 dB, the relative improvements are 71.93\% over MFCC, 81.19\% over FM, 73.49\% over LI, 8.18\% over CMVN. When compared with
other 2D psychoacoustic filters, the Adaptive 2D filter manages to obtain very promising improvements. At Avg 0-20, the relative improvements are 1.08\% over TW-2D and
0.69\% over TFW-2D respectively. For SNR of -5 dB, the relative improvements are 19.60\% over TW-2D and 5.18\% over TFW-2D respectively.

\section{Conclusion}\label{sc:2D_conclusion}
We propose a hybrid feature extraction algorithm based on MFCCs, which successfully implements FM, LI and TI with a simple 2D psychoacoustic filter. This method manages to reflect the asymmetrical nature of the human auditory system. The key feature of the proposed algorithm is that we incorporate an adaptive scheme, which better reflects the frequency-dependent property of masking effects. The speech spectrum is divided into multiple bands. Different psychoacoustic filters are designed to better fit the specific frequency band. 

Moreover, the proposed method does not need any additional training process, making the computational burden very low.
Also, due to the simplicity of the proposed algorithm, it can be easily combined with other algorithms. Another important contribution of this paper is the double transform analysis technique, which enables quantitative analysis of the performance of time-frequency domain filters for different noise types. In particular, we successfully explained the performance difference between the Airport test subset result and the Exhibition test subset result.
Extensive comparison is made against state-of-the-art ASR algorithms based on the AURORA2 database.
Statistically significant improvements are achieved as manifested in the experimental results. 

\section{Appendices}
\subsection{2D Psychoacoustic Filters}\label{ap:2dfilter}
Table \ref{tab:atfw2Dlow} and Table \ref{tab:tfw2Dhigh} give the detailed parameters of the proposed low band and high band 2D psychoacoustic filters.

\begin{table}[!ht]
\small
\renewcommand{\arraystretch}{1.2}
\caption{Temporal Frequency Warped 2D Psychoacoustic Filter (low
band)} \label{tab:atfw2Dlow} \centering
    \begin{tabular}[t]{|c|ccc ccc ccc|}
    \hline
    Freq$\setminus$T & 0 & 1 & 2 & 3 & 4 & 5 & 6 & 7 & 8\\
    \hline
    -1 & -0.0137 &-0.0065 &-0.005  &-0.0041 &-0.0034 &-0.0029  & -0.0025&-0.0022 & -0.0019\\
    0  & 1+$\alpha_{TI}^{low}$       &-0.4736 &-0.3622 &-0.2971 &-0.2508 &-0.215 & -0.1857&-0.1609 & -0.1395 \\
    1  & -0.0914 &-0.0433 &-0.0331 &-0.0272 &-0.0229 &-0.0196  & -0.017 &-0.0147 & -0.0127 \\
    2  & -0.1757 &-0.0832 &-0.0636 &-0.0522 &-0.0441 &-0.0378  & -0.0326&-0.0283 & -0.0245\\
    3  & -0.2386 &-0.113  &-0.0864 &-0.0709 &-0.0598 &-0.0513 & -0.0443&-0.0384 & -0.0333 \\
    4  & -0.2129 &-0.1008 &-0.0771 &-0.0632 &-0.0534 &-0.0458 & -0.0395 &-0.0343 & -0.0297 \\
    5  & -0.0986 &-0.0467 &-0.0357 &-0.0293 &-0.0247 &-0.0212  & -0.0183&-0.0159 & -0.0138\\
    \hline
    Freq$\setminus$T  & 9 & 10 & 11 & 12 & 13 & 14 & 15 & 16 &\\
    \hline
    -1 & -0.0017 &-0.0014 &-0.0012 &-0.001  &-0.0008&-0.0007 &-0.0005 &-0.0004 & \\
    0  & -0.1205 &-0.1036 &-0.0883 &-0.0743 &-0.0614&-0.0495 &-0.0384 &-0.0281 & \\
    1  & -0.011  &-0.0095 &-0.0081 &-0.0068 &-0.0056&-0.0045 &-0.0035 &-0.0026 &  \\
    2  & -0.0212 &-0.0182 &-0.0155 &-0.0131 &-0.0108&-0.0087 &-0.0068 &-0.0049 & \\
    3  & -0.0288 &-0.0247 &-0.0211 &-0.0177 &-0.0147&-0.0118 &-0.0092 &-0.0067 & \\
    4  & -0.0257 &-0.0221 &-0.0188  &-0.0158 &-0.0131&-0.0105 &-0.0082 &-0.0060 &\\
    5  & -0.0119 &-0.0102 &-0.0087   &-0.0073 &-0.0061&-0.0049 &-0.0038 &-0.0028 &\\
    \hline
    \end{tabular}
\end{table}

\begin{table}[!ht]
\small
\renewcommand{\arraystretch}{1.2}
\caption{Temporal Frequency Warped 2D Psychoacoustic Filter}
\label{tab:tfw2Dhigh} \centering
    \begin{tabular}[t]{|c|ccc ccc ccc|}
    \hline
    Freq$\setminus$T & 0 & 1 & 2 & 3 & 4 & 5 & 6 & 7 & 8\\
    \hline
    -1  &   -0.0137 &   -0.0060 &   -0.0046 &   -0.0037 &   -0.0031 &   -0.0026  &   -0.0023&   -0.0019 &   -0.0017 \\
    0   &   1+$\alpha_{TI}^{high}$   &   -0.4375 &   -0.3321 &   -0.2705 &   -0.2268 &   -0.1929 &   -0.1651&   -0.1417 &   -0.1214 \\
    1   &   -0.0914 &   -0.0400 &   -0.0304 &   -0.0247 &   -0.0207 &   -0.0176  &   -0.0151&   -0.0130 &   -0.0111 \\
    2   &   -0.1757 &   -0.0769 &   -0.0584 &   -0.0475 &   -0.0398 &   -0.0339 &   -0.0290&   -0.0249 &   -0.0213  \\
    3   &   -0.2386 &   -0.1044 &   -0.0792 &   -0.0645 &   -0.0541 &   -0.0460 &   -0.0394&   -0.0338 &   -0.0290 \\
    4   &   -0.2129 &   -0.0931 &   -0.0707 &   -0.0576 &   -0.0483 &   -0.0411  &   -0.0352&   -0.0302 &   -0.0258\\
    5   &   -0.0986 &   -0.0431 &   -0.0327 &   -0.0267 &   -0.0224 &   -0.0190  &   -0.0163&   -0.0140 &   -0.0120\\
    \hline
    Freq$\setminus$T  & 9 & 10 & 11 & 12 & 13 & 14 & 15 & 16 &\\
    \hline
    -1  &   -0.0014 &   -0.0012 &   -0.0010  &   -0.0008 &   -0.0007&   -0.0005 &   -0.0004 &   -0.0002 &\\
    0    &   -0.1035 &   -0.0875 &   -0.0730  &   -0.0598 &   -0.0476&   -0.0364 &   -0.0259 &   -0.0161 &  \\
    1   &   -0.0095 &   -0.0080 &   -0.0067 &   -0.0055 &   -0.0044&   -0.0033 &   -0.0024 &   -0.0015 & \\
    2   &   -0.0182 &   -0.0154 &   -0.0128 &   -0.0105 &   -0.0084&   -0.0064 &   -0.0045 &   -0.0028 &\\
    3    &   -0.0247 &   -0.0209 &   -0.0174 &   -0.0143 &   -0.0114&   -0.0087 &   -0.0062 &   -0.0038 &  \\
    4    &   -0.0220 &   -0.0186 &   -0.0155 &   -0.0127 &   -0.0101&   -0.0077 &   -0.0055 &   -0.0034 &   \\
    5    &   -0.0102 &   -0.0086 &   -0.0072  &   -0.0059 &   -0.0047&   -0.0036 &   -0.0026 &   -0.0016 & \\
    \hline
    \end{tabular}
\end{table}

\bibliographystyle{IEEEtranS}
\bibliography{IEEEfull,spenh_dp}

\end{document}